\documentclass[a4paper,11pt,twocolumn,twoside]{article}
\usepackage[utf8]{inputenc}
\usepackage{graphicx}
\usepackage{sepln_en}
\usepackage{fullname}

\usepackage[basque,spanish,main=english]{babel} 

\usepackage{enumitem}
\usepackage{tabularray}
\UseTblrLibrary{booktabs}
\usepackage{tcolorbox}
\tcbuselibrary{skins}
\tcbuselibrary{raster}

\usepackage{url}

\newcommand{\pop}[1]{\underline{\textbf{#1}}}

\title{Evaluating the Evaluator: Summarization Metrics and LLM-Judges beyond English}

\author {Jeremy Barnes$^1$, Naiara Perez$^1$, Alba Bonet$^2$, Begoña Altuna$^3$ \\
$^1$HiTZ Center, University of the Basque Country (UPV/EHU)\\
$^2$Department of Software and Computing Systems, University of Alicante, Spain\\
$^3$GOI Institute, Basque Summer University (UEU)\\
 \url{{jeremy.barnes,naiara.perez}@ehu.eus},\\ \url{alba.bonet@ua.es}, \url{begona.altuna@ueu.eus}
}

\seplntranstitle{Evaluando al evaluador:\\métricas de resumen y LLMs como juez más allá del inglés}

\seplnclave{resumen, metaevaluación, idioma de pocos recursos}

\seplnresumen{El resumen automático de textos se basa en evaluaciones automáticas para determinar la calidad de los modelos de resumen mediante métricas automáticas y modelos LLM-as-a-Judge. Sin embargo, estas técnicas requieren metaevaluación para garantizar que capturen correctamente los juicios humanos. En este artículo exploramos esta metaevaluación más allá del inglés mediante un nuevo corpus multilingüe para la metaevaluación de resúmenes (BASSE), que incluye juicios humanos sobre 2.040 resúmenes abstractivos generados manualmente o por cinco modelos de lenguaje de gran tamaño (LLMs) con cuatro \textit{prompts} diferentes. Para cada resumen, los anotadores evalúan cinco criterios en una escala Likert de 5 puntos: \textit{coherence}, \textit{consistency}, \textit{fluency}, \textit{relevance} y \textit{5W1H}. Luego evaluamos métricas automáticas de resumen y modelos LLM-as-a-Judge. Nuestros resultados muestran que los LLMs propietarios usados como jueces presentan la mayor correlación con los juicios humanos, seguidos por métricas automáticas específicas por criterio, mientras que los LLMs de código abierto usados como jueces presentan bajo rendimiento.}

\seplnkey{summarization, meta-evaluation, under-resourced languages}

\seplnabstract{Automatic text summarization relies on automatic evaluation to quickly determine the quality of summarization models via automatic metrics and LLM-as-a-Judge models. However, these techniques require meta-evaluation to ensure that they capture human judgments correctly. In this paper, we explore this meta-evaluation beyond English by generating a new multilingual summary meta-evaluation dataset (BASSE), which comprises human judgments on 2,040 abstractive summaries, generated either manually or by five Large Language Models (LLMs) with four different prompts. For each summary, annotators evaluate five criteria on a 5-point Likert scale: \textit{coherence}, \textit{consistency}, \textit{fluency}, \textit{relevance}, and \textit{5W1H}. We then benchmark automatic summarization metrics and LLM-as-a-Judge models. Our results show that currently proprietary judge LLMs have the highest correlation with human judgments, followed by criteria-specific automatic metrics, while open-sourced judge LLMs perform poorly.}

\firstpageno{1}

\begin{document}


\setlength\titlebox{20cm} 

\label{firstpage} \maketitle

\section{Introduction}

Automatic text summarization seeks to create a concise and fluent text that captures the main ideas from one or more documents \cite{giarelis2023abstractive}. This task has been studied in depth in Natural Language Processing (NLP) \cite{Dang2006OverviewOD,Dang2008OverviewOD} and plays an important role in mitigating excess information in an age of increasing information glut \cite{navas2012acceso}.

Current approaches to summarization generally rely on abstractive methods, leading to new challenges for evaluation, as abstractive models are more prone to hallucinating information \cite{maynez-etal-2020-faithfulness,laban-etal-2022-summac}.
Evaluation of automatic summaries relies on three main strategies: \textit{i)} human evaluation, \textit{ii)} evaluation using reference summaries and automatic metrics, and \textit{iii)} evaluation via Large Language Models (LLMs), i.e., LLM-as-a-Judge. Human evaluation on certain criteria, such as coherence or fluency, allows for high-quality evaluation of abstractive summaries. However, this approach is not apt for iterating over models, where instead, automatic metrics, e.g., ROUGE \cite{lin-2004-rouge} and, more recently, judge LLMs \cite{leiter-etal-2023-eval4nlp,kim-etal-2024-prometheus} are preferred. 

However, selecting appropriate metrics or judge LLMs relies on meta-evaluations that measure the correlation between human evaluations and automatic evaluation methods \cite{bhandari-etal-2020-evaluating,fabbri-etal-2021-summeval,deutsch-etal-2022-examining}. While these studies provide invaluable information, they have only been performed comprehensively for English, leaving the assessment in other languages to be explored.

We address this gap by collecting human annotations for automatic summaries in Basque and Spanish---languages with respective fragmentary and medium coverage of language resources \cite{Giagkou2023ch4}. 

Specifically, we collect abstractive summaries generated by five LLMs and four different prompts on 45 news articles in each language. We then collect human judgments for each summary with respect to five criteria---namely, \textit{coherence}, \textit{consistency}, \textit{fluency}, \textit{relevance}, and \textit{5W1H}. With this novel dataset, \textbf{BA}sque and \textbf{S}panish \textbf{S}ummarization \textbf{E}valuation (BASSE),\footnote{Data and code available at \url{https://github.com/hitz-zentroa/summarization}} we address the following research questions:

\begin{itemize}[leftmargin=*]
    \item \textbf{RQ1:} Do automatic metrics for evaluating summarization correlate well with human judgments?
    \item \textbf{RQ2:} Can judge LLMs replicate human judgments for summarization?
    \item \textbf{RQ3:} Which summary criteria (\textit{coherence}, \textit{consistency}, \textit{fluency}, \textit{relevance}, and \textit{5W1H}) do state-of-the-art LLMs struggle with in Basque and Spanish?
\end{itemize}

Our results indicate that proprietary LLMs-as-a-Judge models currently have the highest correlation with human judgments on our data, while several traditional metrics also correlate well with specific criteria. On the contrary, open-source LLMs-as-a-Judge currently do not correlate well with human judgments on our data.

Our contributions are hence the following: \textbf{(1)} We present a new multilingual corpus (Basque, Spanish) of summaries generated by five LLMs  manually annotated for five criteria on a 5-point Likert scale; \textbf{(2)} we study the correlation between human judgments with the results of automatic evaluation metrics and LLM judges in languages other than English; and \textbf{(3)} we additionally provide the first corpus for summarization in Basque, consisting of 22,525 news articles and their subheads.

\section{Related Work}
\label{sec:related-work}

\paragraph{Summarization Evaluation Methods.}

Manual evaluation becomes an impossible task as corpus size increases, motivating the need for automatic metrics for summarization evaluation, e.g., ROUGE variants \cite{lin-2004-rouge,rankel-etal-2013-decade,graham-2015-evaluating}, S3 \cite{peyrard-etal-2017-learning}. More recently, using LLMs to automatically evaluate summaries---LLMs-as-a-Judge---has shown promise \cite{leiter-etal-2023-eval4nlp,kim-etal-2024-prometheus,lyu2025exploringlimitoutcomereward}. These models often show higher correlations with human judgments than previous token-based metrics. However, nearly all of these metrics and models have been designed for English, and their reliability on non-English summaries remains insufficiently understood. Recent work has raised concerns about the reliability of LLM judges, showing performance variability across languages and limited agreement with human judgements \cite{son2025mmevalm,fu-liu-2025-reliable,ponce-etal-2026-judging}.

\paragraph{Meta-evaluation of Automatic Metrics.}

Meta-evaluation is necessary to establish the reliability of automatic metrics, and is typically done by studying the correlation between metric scores and human judgments \cite{lin-2004-rouge}. \namecite{fabbri-etal-2021-summeval}, for example, collect human evaluations for 23 English summarization models over four dimensions (\textit{coherence}, \textit{consistency}, \textit{fluency}, and \textit{relevance}) and find that automatic metrics generally correlate poorly with the human annotations. While some studies have examined meta-evaluation beyond English \cite{saggion-etal-2010-multilingual,clark-etal-2023-seahorse,koto-etal-2021-evaluating}, none of them apply a range of automatic metrics or LLM judges and most use automatically extracted subhead-article pairs as a proxy for reference summaries, limiting the reliability of the findings.

\section{Methodology}
\label{sec:methodology}

\begin{table*}[ht]
    \centering
    \begin{tblr}{
       colspec={*{11}{r}},
       cells={font=\small},
       columns={rightsep=5pt},
       row{1,2}={c,m,font=\small\bfseries},
       column{1}={l,font=\small\bfseries},
       column{1,6}={rightsep=10pt},
       rowsep=0pt
    }
    \toprule
                  & \SetCell[c=5]{c} Basque & & & & & \SetCell[c=5]{c} Spanish \\
                   \cmidrule[lr]{2-6} \cmidrule[lr]{7-11}
                  & Doc & ¬ eu & Tok & Voc & {s/d}           &  Doc & ¬ es &  Tok & Voc & {s/d} \\
    \midrule
        Source article   &  45 &   - & 22,862 & 7,328 & 34 &  45 &   - & 38,551 & 8,200 & 30 \\
    \midrule
        Subhead   &  45 &   0 &   1,562 &    927 &  3 &  45 &   0 &   1,040 &    578 & 1 \\
        Human summary     &  75 &   0 &   8,867 &  3,428 &  6 &  75 &   0 &  11,635 &  3,180 & 5 \\
        LLM summary      & 900 &  54 & 133,755 & 15,755 & 10 & 900 &  27 & 158,945 & 10,844 & 8 \\
    \bottomrule
\end{tblr}
    \caption{Statistics of the BASSE corpus, including the source articles and their summaries. We report the number of documents (Doc)---and, of those, how many are not in the target language (¬ eu or ¬ es)---, total tokens (Tok), vocabulary size (Voc), and sentences per document (s/d). Splitting and tokenization were done with \texttt{stanza} \cite{qi-etal-2020-stanza} and language identification with \texttt{lingua-py}.\protect\footnotemark[10]}
    \label{tab:basse}
\end{table*}

We develop our multilingual summarization evaluation dataset through three main steps. First, we collect news articles from quality sources in both target languages ($\S$\ref{ssec:data-collection}). Second, we generate a battery of summaries that includes original subheads, human-written summaries, and automatic summaries from various LLMs using different prompts ($\S$\ref{ssec:summary-collection}). Finally, we conduct expert annotation using five criteria through a multi-round process for reliable assessment ($\S$\ref{ssec:human-annotations}).

\subsection{Data Collection}
\label{ssec:data-collection}

Data collection was conducted with the aim of collecting highly representative data. We randomly collected news articles covering diverse topics and genres to ensure variety across both languages. For Basque, we collect 45 articles from Berria,\footnote{\url{https://www.berria.eus}} the main newspaper in this language.\footnote{In addition to the 45 articles we annotate, we also collect 22,525 Basque articles from Berria with their subhead which we release with BASSE.}
For Spanish, another set of 45 news articles was randomly extracted from the test set of the MLSUM dataset \cite{scialom-etal-2020-mlsum}. The Spanish subset of MLSUM contains news articles of El País newspaper\footnote{\url{https://www.elpais.com}} and provides (news body, subhead, URL) triplets. 

\subsection{Summary Collection}
\label{ssec:summary-collection}

For each of the 45 articles in both languages, we collected three types of summaries, the final multilingual dataset comprising 90 source articles and 2,040 summaries:

\begin{itemize}
    \item \textbf{Subhead}: We extracted the original article subheads to serve as baseline summaries \cite{moritz-2015-cnn,narayan-etal-2018-dont}.
    \item \textbf{Human}:  We collected manually written summaries from our expert annotators. For the first 15 articles used to compute agreement ($\S$\ref{sssec:annotation-procedure}), each article has three human summaries. The remaining 30 articles were divided among the annotators, with one human summary each, yielding a total of 75 manual summaries per language.
    \item \textbf{Automatic}:  We generated automatic summaries using five different LLMs with four different prompts, leading to 900 automatic summaries per language. The specific models and prompting strategies are described in the following subsections.
\end{itemize}

\subsubsection{Models}
\label{sssec:models}

We selected a number of proprietary and open-source LLMs that are able to provide summaries in Basque and Spanish. Proprietary models included Anthropic's Claude,\footnote{\url{https://www.anthropic.com/claude}} OpenAI's GPT-4o,\footnote{\url{https://openai.com/index/gpt-4o-system-card}} and Reka AI's Reka \cite{team2024reka}.\footnote{We used Claude 3.5 Sonnet for Rounds 1, 2, and 3 and Claude 3.5 Haiku for Round 4. Similarly, we used Reka-Core for Rounds 1,2, and 3 and Reka-Flash for Round 4.} The open-source options were Cohere's Command R+\footnote{\url{https://docs.cohere.com/v2/docs/command-r-plus}} and Meta's Llama 3.1 70B Instruct \cite{dubey2024llama}. All models were prompted using their default configurations through their respective native chat interfaces, with the exception of Llama, which was accessed via HuggingFace's chat interface\footnote{\url{https://huggingface.co/chat}} due to the absence of a proprietary chat platform.

\subsubsection{Prompts}
\label{sssec:prompts}

Along with the selection of those five different models, we experimented with four prompts to collect summaries from the LLMs. These prompts, presented below, were designed in order to obtain fairly diverse summaries. Although multilingual LLMs often show stronger performance when prompted in English \cite{Shi2022LanguageMA,etxaniz-etal-2024-multilingual}, in early experiments, we found that prompting the LLMs in English led to the summaries also being given in English. Therefore, we opted to prompt the model directly in the language in which it should write the summary. Table \ref{tab:prompts} in Appendix \ref{ann:prompts} shows each prompt in Basque and Spanish, as well as its translation to English; their descriptions are provided below.

\begin{itemize}
    \item \textbf{Base}: This prompt asks the LLM to summarize the following document.
    \item \textbf{Core}: This prompt asks the model to think about the most important information in the original document and use that information to create a summary for the following document.
    \item \textbf{5W1H}: This prompt, which is a further derivation of Core, uses an approach common in journalism \cite{zhang2019dynamic,chakma20205W1H} consisting of answering six key questions: the five ``W'' questions (who, what, when, where, why) and the 1 ``H'' question (how). 
    \item \textbf{Too long, didn't read (tldr)}: This prompt appends the internet shorthand to elicit a concise summary.
\end{itemize}

\footnotetext[10]{\url{https://github.com/pemistahl/lingua-py}}\addtocounter{footnote}{1}

\subsection{Human Annotation}
\label{ssec:human-annotations}

We followed the annotation procedure of \namecite{fabbri-etal-2021-summeval}, who based their criteria largely on \namecite{kryscinski-etal-2019-neural} and \namecite{Dang2005OverviewOD}. For our experiment, besides the following descriptions of the used criteria, we elaborated a detailed set of annotation guidelines, which we provide in Appendix \ref{ann:guidelines}, while Appendix \ref{ann:examples} contains real examples of annotated summaries in Basque and Spanish.

\subsubsection{Criteria}
\label{sssec:criteria}

We maintain the core criteria of previous work \cite{fabbri-etal-2021-summeval}---\textit{coherence}, \textit{consistency}, \textit{fluency}, and \textit{relevance}---and extended the framework with an additional criterion, \textit{5W1H}, which measures how well important information is preserved from the source by following a journalistic approach. It evaluates whether key content is presented completely by addressing the six fundamental questions (see prompt 5W1H in Appendix \ref{ann:prompts}). The summary should not lack any important information available in the source document. This recall-oriented criterion corresponds roughly to \textit{coverage} in   \namecite{Koto2020FFCIAF}, although it differs crucially in that we calculate \textit{5W1H} relative to the original document, whereas they use the reference summary to calculate \textit{coverage}.

\begin{table*}[htb]
    \centering
    \begin{tblr}{
   colspec={*{12}{c}},
   cells={font=\small},
   columns={rightsep=5pt},
   row{1,2}={c,font=\small\bfseries},
   column{1}={c,font=\small\bfseries},
   column{1,2,7}={rightsep=10pt},
   rowsep=0pt
}
\toprule
    &      & \SetCell[c=5]{c} Basque & & & &  & \SetCell[c=5]{c} Spanish \\
\cmidrule[lr]{3-7} \cmidrule[lr]{8-12}
R\# & Doc & Coh & Con & Flu & Rel & 5W1H & Coh & Con & Flu & Rel & 5W1H \\
\midrule
1   & 210 & 0.39 & 0.56 & 0.68 & 0.34 & 0.55 & 0.31 & 0.18 & 0.13 & 0.22 & 0.39 \\
2   & 210 & 0.59 & 0.63 & 0.75 & 0.54 & 0.64 & 0.66 & 0.37 & 0.35 & 0.49 & 0.58 \\
3   & 105 & 0.66 & 0.44 & 0.70 & 0.63 & 0.72 & 0.29 & 0.19 & 0.34 & 0.20 & 0.39 \\
\bottomrule
\end{tblr}%
    \caption{Ordinal Krippendorff's $\alpha$ across triple-annotated rounds (R\#). Rounds 1 and 2 were conducted on the same set of summaries (Doc), before and after guideline discussion respectively, while round 3 involved a new set of summaries.}
    \label{tab:alphas}
\end{table*}

\subsubsection{Annotation Procedure}
\label{sssec:annotation-procedure}

We performed three rounds of expert annotation, where the first two served to refine the annotation guidelines. Given the generally poor agreement of crowdsourced annotations for summarization  \cite{gillick-liu-2010-non,fabbri-etal-2021-summeval}, the annotators were recruited among NLP linguists and engineers who had extensive experience with annotation projects and knowledge of summarization.

In the first round, three annotators evaluated summaries from 10 articles in each language, with each (model, prompt, document) combination being annotated for all five criteria. They additionally annotated the same criteria for the articles' subheads as baseline summaries and provided their own summaries for each text (which were evaluated by the other two annotators). After calculating agreement scores, annotators discussed the cases of disagreement and refined the guidelines. In the second round, the same set of summaries was re-annotated following the refined guidelines. The third round involved a new set of 5 articles, with all summaries being triple-annotated. Following the third round, given the acceptable agreement levels achieved, annotators proceeded to individually annotate an additional set of 10 documents each, which brings the total to 45 annotated articles.

After each round of annotation, we calculated inter-annotator agreement using ordinal Krippendorff's $\alpha$  \cite{krippendorff2013content} to compare three-way agreement for each of the criteria, and additionally calculated quadratic Cohen's $\kappa$  \cite{cohen1960coefficient} for pairwise agreement. 

\section{The BASSE Corpus}
\label{sec:corpus}

We now present BASSE, the manually annotated multilingual corpus of 2,040 news summaries in Basque and Spanish, analyzing inter-annotator agreement levels, corpus statistics, and qualitative differences across summarization approaches. The diversity of summary generation methods in BASSE enables a robust assessment of summary evaluation metrics under varied conditions.

\subsection{Inter-annotator Agreement}
\label{ssec:corpus-iia}

Initial annotations (R\#1) yielded moderate agreement levels for Basque across criteria and lower agreement for Spanish (Krippendorff's $\alpha$ correlations for each round and language are shown in Table \ref{tab:alphas}.). By the final round (R\#3), Basque annotations maintained acceptable agreement levels (0.44--0.72), with Spanish remaining at lower values (0.19--0.39). Consistent with prior work \cite{falke-etal-2019-ranking,kryscinski-etal-2020-evaluating}, \textit{consistency} proved to be the most challenging for annotators to align on---0.44 and 0.19 for Basque and Spanish, respectively, in the final round. This challenge can be attributed to the inherent complexity of evaluating factual alignment, where annotator's interpretations of source article details sometimes diverge.

While pairwise quadratic Cohen's $\kappa$ values follow similar patterns to the Krippendorff's $\alpha$ scores, raw agreement percentages show moderate to high concordance in several categories for Spanish, notably \textit{fluency} (83--90\%) and \textit{consistency} (65--67\% (See Appendix  \ref{ann:iia}). This apparent contradiction is explained by the score distributions, where Basque summaries exhibit a broader distribution across the quality spectrum, while Spanish summaries tend to concentrate in the higher range of the quality scale, leading to higher expected agreement and consequently lower chance-corrected scores, a known problem with IAA scores \cite{Xu-2014-kappa}.

Note that our agreement scores, including those for Spanish annotations, are similar to or exceed those reported in previous large-scale summary evaluation studies of reference \cite{fabbri-etal-2021-summeval,clark-etal-2023-seahorse,aharoni-etal-2023-multilingual}.

\begin{figure*}[htb]
    \centering
    \includegraphics[width=\textwidth,trim={6pt 8pt 6pt 8pt},clip]{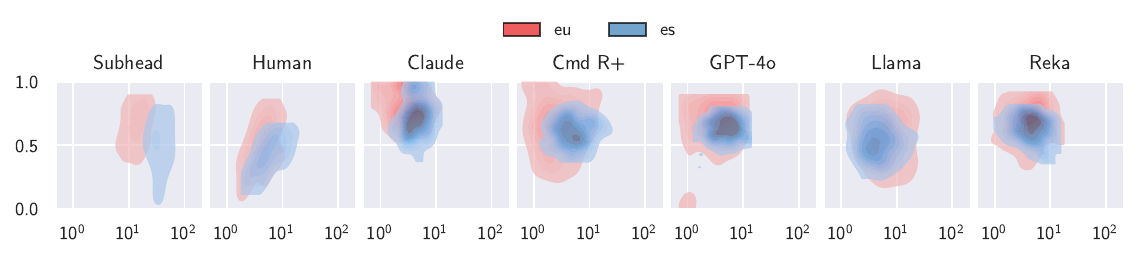}
    \caption{Distribution of compression ratios (\texttt{x} axis) and novel 2-grams (\texttt{y} axis) for different summarization approaches, in Basque (eu) and Spanish (es). Higher compression values ($\rightarrow$) indicate more concise summaries, while higher novel 2-gram values ($\uparrow$) suggest more abstractive summarization.}
    \label{fig:data-stats} 
\end{figure*}

\subsection{Quantitative Analysis}
\label{ssec:corpus-quantitative}

Table \ref{tab:basse} presents the statistics of our corpus, which comprises 45 articles with their subheads, 75 human-written summaries, and 180 summaries from each model. The source articles contain between 20--40K tokens in total, with Basque showing a markedly lower token count due to its agglutinative nature. 

A small number of the automatically generated summaries are not written in the target languages Basque and Spanish (6\% and 3\% respectively). Summary vocabulary sizes range between 3K and 10K tokens, and they vary considerably in length and richness across different approaches. Statistics for LLM summaries broken down by author model and prompt can be consulted in Appendix \ref{ann:basse-stats}, Table \ref{tab:basse-detail}.

In order to quantify the diversity of summarization strategies, we evaluate them through two metrics: \textit{compression ratio} and \textit{novel 2-gram}. Compression ratio indicates summary conciseness with respect to the source article \cite{segarra-soriano-etal-2022-dacsa}, while novel 2-grams measure the percentage of new word pairs in the summary \cite{kryscinski-etal-2018-improving}. The trade-off between these two is shown Figure \ref{fig:data-stats}.

Subheads achieve the highest compression (10--100$\times$), while human and model summaries typically compress around 6$\times$, with some notable exceptions. GPT-4o occasionally produced expansive content matching or exceeding source length, while Llama consistently generated more compressed summaries.

The novel 2-gram analysis shows that model-generated summaries are more abstractive than human-written ones, particularly from Claude (though this finding is partially influenced by Claude's outputs in languages other than the target one). In contrast, GPT-4o sometimes opted for a more extractive approach, closely reproducing the original text as mentioned earlier.

\subsection{Qualitative Analysis}
\label{ssec:corpus-qualitative}

The manual evaluation of the summaries (see Table \ref{tab:human-eval-average} and detailed results per model and prompt combination in Appendix \ref{ann:summarization-results}) shows large quality differences across summarization approaches. While human-written summaries achieve the highest average scores (4.75 for Basque, 4.68 for Spanish), the subheads fail at covering key information (\textit{5W1H}). 

Among the model-generated summaries, we observe a clear linguistic divide, with Spanish summaries generally achieving higher \textit{fluency} scores. This gap is most pronounced for Command R+, where poor \textit{fluency} in Basque (2.64 compared to 4.92 in Spanish) consequently impacts its \textit{consistency}. GPT-4o shows the best performance in both languages, although all models show varying performance patterns across prompting strategies.

Most LLMs achieve \textit{consistency} scores comparable to humans, and some models occasionally surpass human performance in \textit{relevance} and \textit{5W1H} coverage. In fact, the 5W1H prompt seems to compel models to provide a better coverage of key information.

All models normally generate summaries with a suitable format. However, Claude shows a tendency to generate bullet-point summaries, which negatively impacts its \textit{coherence} scores, and occasionally produces English text with the tldr prompt, as noted earlier. Additionally, it is the most prone to generating introductory sentences for the summaries, e.g., ``\textit{Aquí está el resumen del texto:}'' (Here is the summary of the text:), which negatively impacts its \textit{relevance} scores. 

\section{Evaluation}
\label{sec:evaluation}

In this section, we calculate the correlation between the human annotations from $\S$\ref{sec:corpus} and a series of evaluation metrics and LLM-as-a-Judge commonly used for evaluating Natural Language Generation (NLG). 
For rounds with more than one annotation per summary, we take the average of the three values for each criterion. We calculate both Spearman's ($\rho$) and Kendall's ($\tau$) rank correlation coefficients at system-level following \namecite{louis-nenkova-2013-automatically} and \namecite{fabbri-etal-2021-summeval}, but report here only Spearman, as the results do not vary depending on the coefficient.

\subsection{Evaluation Metrics}
\label{ssec:evaluation-metrics}

We select a subset of metrics used in SummEval \cite{fabbri-etal-2021-summeval} which require minimal changes to work for languages other than English, specifically \textbf{ROUGE} \cite{lin-2004-rouge}, \textbf{ROUGE-we} \cite{ng-abrecht-2015-better}, \textbf{BertScore} \cite{Zhang2019BERTScoreET}, \textbf{BLEU} \cite{papineni-etal-2002-bleu}, \textbf{CHRF} \cite{popovic-2017-chrf}, \textbf{METEOR} \cite{banerjee-lavie-2005-meteor}, \textbf{CIDEr} \cite{vedantam2015-cider}, and \textbf{Data Statistics} \cite{grusky-etal-2018-newsroom}. 

Due to their dependence on specialized English training data or English-only models, several of the original metrics are excluded, namely, BLANC \cite{vasilyev-etal-2020-fill}, S3 \cite{peyrard-etal-2017-learning}, SummaQA \cite{scialom-etal-2019-answers}, MoverScore \cite{zhao-etal-2019-moverscore}, Sentence Mover's Similarity \cite{clark-etal-2019-sentence}, and SUPERT \cite{gao-etal-2020-supert}. 

For several other metrics (i.a., ROUGE-we, BertScore, METEOR, CIDEr, and Data Statistics), we implement the necessary changes to allow for evaluation in Spanish or Basque. Before passing the candidate or reference summaries to each metric, we perform metric specific preprocessing of the texts (see further details in Appendix \ref{ann:metrics}). 

\subsection{LLM-as-a-Judge}
\label{ssec:judges}

We further experiment with using LLM-as-a-Judge systems to predict the five criteria we use for human evaluation. Currently, no LLM-judge models officially support Basque or Spanish evaluation specifically. Therefore, we selected a range of models that have demonstrated strong performance as judges in English and that offer some multilingual capabilities. Our selection represents a diverse set of characteristics: open-source versus proprietary models, and varying parameter sizes and specializations. 

Following these selection principles, we evaluate nine models as judges: \textbf{Prometheus2 7B} and \textbf{8x7B} \cite{kim-etal-2024-prometheus}, \textbf{M-Prometheus 7B} and \textbf{14B} \cite{pombal2025mprometheussuiteopenmultilingual}, \textbf{Selene Mini} \cite{alexandru2025atlaseleneminigeneral}, \textbf{Qwen2.5 7B} and \textbf{72B} \cite{qwen2025qwen25technicalreport}, \textbf{GPT-4o mini}\footnote{\texttt{gpt-4o-mini-2024-07-18}} and \textbf{GPT-4o}.\footnote{\texttt{gpt-4o-2024-08-06}}

We use the codebase from Prometheus-Eval\footnote{\url{https://github.com/prometheus-eval/prometheus-eval}} and format the prompt to include the original instruction, the annotation rubric for the criteria, the corresponding human-generated reference answer, and the summary to be evaluated. We evaluate only one criterion at a time. Although the original documents and summaries are in either Basque or Spanish, we use English for the instructions and annotation rubric, as preliminary results indicated that several models struggle to follow the directions otherwise. The full prompt template used for all judge models is shown in Appendix \ref{ann:prompts}, Figure \ref{fig:prometheus}.

\section{Results}
\label{sec:results}

\begin{figure*}[t]
    \centering
    \includegraphics[width=\textwidth,trim={6pt 0 6pt 0},clip]{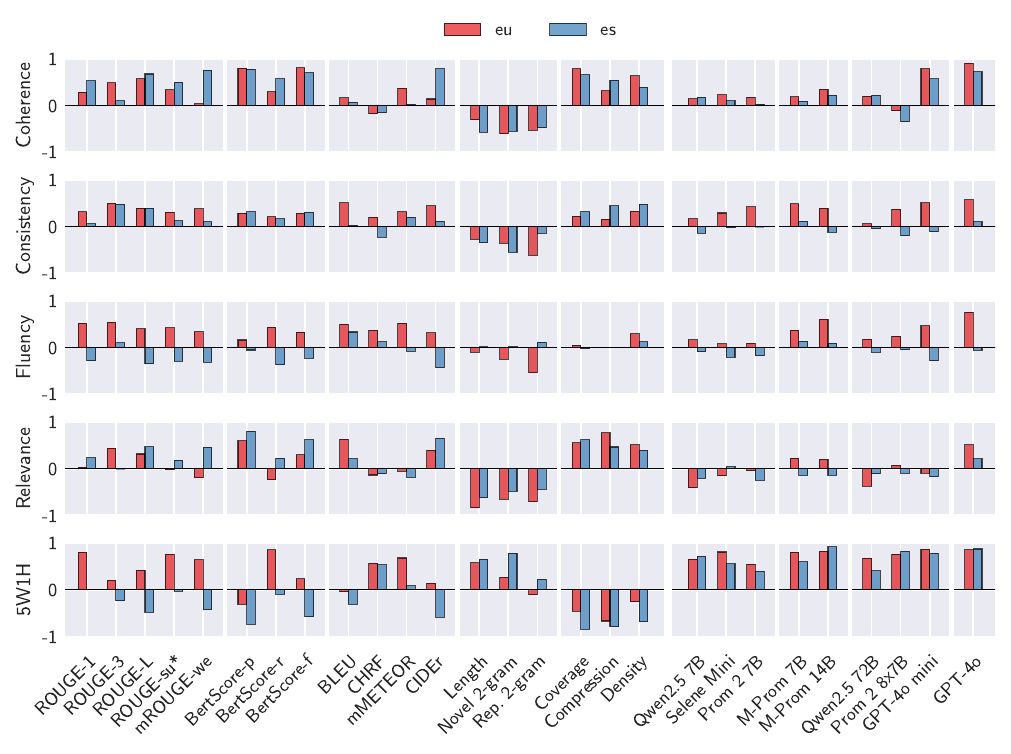}
    \caption{Spearman's model-level rank correlation between automatic metrics (left) or LLM judges (right), and human annotations, in Basque (eu) and Spanish (es).}
    \label{fig:spearman-eu-es} 
\end{figure*}

In this section, we answer the research questions by analyzing Spearman's $\rho$ correlations between model rankings derived from automatic evaluations and those from human judgments. Selected results are shown in Figure \ref{fig:spearman-eu-es}, while full results of both Spearman's $\rho$ and Kendall's $\tau$ rank correlation coefficients can be consulted in Appendix \ref{ann:correlation-evaluation}. 

\subsection*{RQ1: Do automatic metrics for evaluating summarization correlate well with human judgments in Basque and Spanish?} 

Overall, we observe notable patterns across languages in metric performance. Traditional metrics show similar trends in both Basque and Spanish for \textit{coherence}, \textit{consistency}, and \textit{relevance}, with several metrics showing strong correlations with human judgments, despite the gaps in annotator agreement for these criteria. Notably, \textit{fluency} and \textit{5W1H}---the criteria with highest human agreement---result in nearly opposite tendencies between languages, with better correlations in Basque.

Automatic metrics have a significant correlation with \textit{coherence} with overall absolute mean Spearman correlations of 0.548 and 0.454, with BertScore-p giving the most consistent results. 

\textit{Consistency} has the lowest absolute mean correlations between metrics and human judgments (0.201/0.257), indicating that the metrics struggle to capture this criterion.

\textit{Fluency} similarly has low overall absolute mean correlations (0.367/0.164). The same metric often gives opposite results in Basque and Spanish, as is the case of most ROUGE variants or BertScore. \textit{Relevance} has relatively high absolute mean correlations (0.501/0.358), where statistics-based metrics have generally high correlations in both languages. Finally, \textit{5W1H} has moderate absolute mean correlations (0.375/0.463). Compression is the single automatic metric that best captures \textit{5W1H} across both languages (-0.665/-0.792), although coverage (-0.848) and novel unigram (0.816) perform the best in Spanish.

In summary, several metrics correlate strongly for both languages ($\rho >$ 0.6): for \textit{coherence} and \textit{relevance}, BertScore-p shows a strong correlation in both languages, and compression performs well for \textit{5W1H}. In Basque, ROUGE-2 and ROUGE-3 capture \textit{fluency} well, while no automatic metric performs well on \textit{fluency} in Spanish, likely due to the low variance in scores. Similarly, there is no metric that consistently captures \textit{consistency} across both languages. Metrics originally designed for machine translation evaluation (namely, BLEU, CHRF, and METEOR) do not stand out for any specific summarization criterion, generally exhibiting weak correlations across most evaluation dimensions in both languages.

\subsection*{RQ2: Do LLM-as-a-Judge models have the ability to replicate human judgments for summarization?} 

In general, proprietary LLM-as-a-Judge models (GPT-4o Mini and GPT-4o) give the highest correlations with human judgments on most criteria. GPT-4o, for example, gives the highest correlation in four of five criteria for Basque (0.512 $\le \rho \le$  0.909) and one of five in Spanish,  (-0.055 $\le \rho \le$ 0.876). The smaller GPT-4o mini performs similarly, -0.097 $\le \rho \le$ 0.874 in Basque and -0.28 $\le \rho \le$ 0.783 in Spanish.

The open-source LLM-as-a-Judge models, on the other hand, do not generally outperform the automatic metrics. As an exception, for \textit{5W1H} the M-Prometheus models achieve high correlations in Basque and Spanish ($\rho \ge$ 0.598) while Selene Mini has $\rho$ of 0.812 in Basque. M-Prometheus 14B has the third highest correlation for fluency in Basque (0.609). Prometheus 2 8x7B, M-Prometheus 14B, and Qwen2.5-72B-Instruct perform as poorly as their smaller versions, suggesting that the low correlation achieved is not due to model size. Furthermore, Prometheus 2 and M-Prometheus judges show more correlation in Basque than in Spanish. 

Summing up, proprietary models show relatively strong correlations with human judgments (GPT-4o above all, and GPT-4o mini to a lesser degree) while open-source models currently are only able to capture \textit{5W1H}.

\subsection*{RQ3: Which summary criteria (\textit{coherence}, \textit{consistency}, \textit{fluency}, \textit{relevance}, \textit{5W1H}) do state-of-the-art LLMs struggle with in Basque and Spanish?} 

As noted earlier, our goal was not to produce optimal summaries but rather to gather a diverse range of summary qualities for meta-evaluation purposes. We deliberately avoided hyperparameter tuning or extensive prompt engineering that may have yielded better results. Thus, the following results better represent what users can expect out of the box from these models.

In Basque, \textit{coherence} is the criterion with the lowest score, with an average of 3.46/5.00 over all model prompt combinations, followed by \textit{relevance} (3.74), while \textit{consistency} (4.29) and \textit{5W1H} (4.06) score slightly better. That is, Basque summaries are fairly faithful to source content and comprehensive in their information coverage, yet they struggle with logical flow and content relevance. The moderate \textit{fluency} average score (3.87) masks considerable variation in linguistic competence between different models. 

In Spanish, the models generally achieve better scores. \textit{Fluency} has the highest score (4.82), followed by \textit{consistency} (4.68). \textit{Coherence} is lowest (3.88) while \textit{relevance} (4.14) and \textit{5W1H} (4.18) lie in the middle. In other words, Spanish summaries excel in linguistic quality and faithfulness while providing mostly complete and relevant information, although logical connectivity between ideas needs some improvement. Detailed results are available in Appendix \ref{ann:summarization-results}.

\section{Discussion}

This work describes an end-to-end methodology for the assessment of the existing summary evaluation metrics and LLM-as-a-Judge models. A key proposal in the annotation and summary quality evaluation steps is the introduction of the \textit{5W1H} evaluation criterion, which has shown promise as a recall-oriented measure of important information from the original document. One may expect it to be the inverse of \textit{relevance}---a precision-oriented measure of information transfer---but our experiments suggest that there is no direct correlation between them ($\rho$ = -0.16 and -0.34 at model level) and different metrics capture each criterion better. In fact, while LLM-as-a-Judge models generally have $\rho >$ 0.561 for \textit{5W1H}, they have close to random correlations for \textit{relevance} (see Figure \ref{fig:spearman-eu-es}, and Tables \ref{tab:correlation-es} and \ref{tab:correlation-eu}). 

Building on the evaluation criteria, we observed that proprietary LLM-as-a-Judge models generally outperform traditional metrics, especially on \textit{coherence} and \textit{5W1H}, while the open-source LLM-as-a-Judge models perform much worse (idem.). This performance gap is likely due to the fact that neither Prometheus 2, Selene Mini nor Qwen 2.5 are trained as judges for languages other than English, while the proprietary models may have such training. M-Prometheus \cite{pombal2025mprometheussuiteopenmultilingual}, despite being trained on 6 languages---which do not include Basque or Spanish---, shows slightly better correlations with humans but still lags behind GPT-4o. Our experiments thus serve as an interesting testing ground for judge LLMs in cross-lingual contexts. Although multilingual models often can transfer knowledge across languages, this capability does not appear to extend to summary evaluation in our study. This does not indicate that multilingual models are not possible, but rather points to the importance of developing truly multilingual judge models to enable high quality evaluation beyond English. 

Finally, we observed that, although LLMs are capable of producing mostly faithful summaries (see \textit{consistency} scores in Table \ref{tab:human-eval-average} and \ref{tab:human-eval-detail}), \textit{consistency} is currently poorly captured by LLM-as-a-Judge models, as well as by all automatic metrics, contrary to previous findings in English \cite{fabbri-etal-2021-summeval}. This is likely due to the choice of summarization models, as previous meta-evaluations have included summaries from extractive summarization models and simpler abstractive models. Current abstractive summarization models often have subtle inconsistencies \cite{kryscinski-etal-2020-evaluating,yang-etal-2024-fizz}, making them more difficult to capture.

\vspace{2cm}
\section{Conclusion}
\label{sec:conclusion}

In this work we try to assess the adequacy of automatic methods to evaluate summaries in languages other than English. As an experimentation ground, we present BASSE, a dataset of 2,040 automatic summaries in Basque and Spanish that we collected and annotated manually across five criteria on a 5-point Likert scale: \textit{coherence}, \textit{consistency}, \textit{fluency}, \textit{relevance} and \textit{5W1H}. The inter-annotator agreement study performed on a subset of triple-annotated summaries yields better scores than those found in prior large-scale summarization evaluation research. We used this dataset to perform a meta-evaluation of traditional automatic metrics and LLM-as-a-Judge models beyond English contexts. Further, we share the first automatically collected summarization dataset in Basque, which comprises 22,525 article-subhead pairs.

We observed a hierarchy of evaluation approaches: proprietary LLM-as-a-Judge models demonstrate the strongest overall correlations with human judgments, particularly for \textit{coherence} and \textit{5W1H} coverage. On the other hand, traditional metrics like BertScore-p, ROUGE variants, and compression ratio show strong criterion-specific performance but lack consistency across all dimensions. At the bottom of this hierarchy, open-source judge LLMs perform notably poorly, suggesting that current multilingual capabilities do not effectively transfer to the task of summarization evaluation in non-English languages.

Among the evaluation criteria, \textit{coherence} and \textit{relevance} were most reliably captured by automatic metrics, with BertScore-p showing strong correlations across both languages. The \textit{5W1H} criterion was consistently assessed by proprietary judge models. \textit{Fluency} exhibited language-dependent results, with several metrics performing well for Basque but struggling with Spanish. \textit{Consistency} proved the most challenging criterion overall, with no metric or model showing strong correlations across both languages, reflecting the difficulty of detecting subtle factual misalignments in today's abstractive summarization systems' output.

We make the dataset and code available to support future research on multilingual evaluation methods for automatic summarization.

\section*{Limitations}

Although our corpus collection and annotation methodology are the same for both Basque and Spanish, the outcome differs between languages. In Basque, the summaries we collected are generally of poorer quality, but the IAA is high, while for Spanish the quality of all summaries is quite good, but the IAA is lower. This seems to be an artifact of the higher quality of the summaries in Spanish. Annotators rarely used the lower end of the scale (1--3) for several criteria, and the expected agreement is therefore higher, affecting IAA scores, despite raw agreement often being over 80\%. 

Beyond language-specific differences, the domain of our study offers important context for interpreting our findings. This meta-evaluation is performed on newswire data, and our results may not necessarily transfer well to other domains or formulations of summarization, as newswire texts tend to be written in a declarative style, avoid negative constructions, and explicitly represent \textit{5W1H} information. However, the standard \textit{coherence}, \textit{consistency}, \textit{fluency} and \textit{relevance} metrics have been successfully used to annotate summaries in other domains, and it is likely that the new \textit{5W1H} metric can be easily adapted to different types of texts according to the relevance of each of the \textit{5W1H} elements in text.

Finally, note that most of the news articles in our dataset have a single human-generated summary, which limits our ability to exploit the variability of human summaries. However, given a set annotation budget and the goal of creating a large and varied set of summaries, previous research has shown it is preferable to increase the number of instances, even if they are single annotated \cite{steen-markert-2021-evaluate}.  

\section*{Acknowledgements}

This work has been partially supported by the European Union under Horizon Europe (Project LUMINOUS, grant number 101135724), the Ministry of Science and Innovation of the Spanish Government (DeepThought project: PID2024-159202OB-C21), the Ministerio para la Transformación Digital y de la Función Pública - Funded by EU – NextGenerationEU within the framework of the project \textit{Desarrollo de Modelos ALIA}, the Basque Government (IKER-GAITU project), and the University of Alicante, the Spanish Ministry of Science and Innovation (MCIN), and the European Regional Development Fund (ERDF) through the project HEART-NLP (PID2024-156263OB-C22); funded by MCIN/AEI/10.13039/501100011033 / “ERDF, EU”.
Also, this work was partially supported by HumanAIze projects: AIA2025-163322-C61 and AIA2025-163322-C63 funded by MICIU/AEI/10.13039/501100011033.

\bibliographystyle{fullname}
\bibliography{custom}

\appendix

\section{Prompts}
\label{ann:prompts}

Table \ref{tab:prompts} shows the 4 prompt variants used to collect LLM summaries in Basque and Spanish, as well as their translation to English. 
Figure \ref{fig:prometheus} contains the prompt passed to judge LLMs, for which we use the codebase from Prometheus-Eval.\footnote{\url{https://github.com/prometheus-eval/prometheus-eval}} We format the prompt to include the original instruction, the annotation rubric for the criteria---one at a time---(see Appendix \ref{ann:guidelines}), a human-generated reference answer, and the summary to be evaluated.

\begin{table*}[htb]
    \centering
    \begin{tblr}{
    width=\textwidth,
    colspec={rXXX},
    cells={l,font=\small},
    columns={rightsep=5pt},
    row{1}={c,font=\small\bfseries},
    column{1}={l,m,font=\small\bfseries},
    rowsep=0pt
}
\toprule
    & {Basque} & {Spanish} & {English translation} \\
\midrule
Base
    & \selectlanguage{basque}Laburtu hurrengo testua.\texttt{\textbackslash n\textbackslash n[doc]}
    & \selectlanguage{spanish}Resume el siguiente texto.\texttt{\textbackslash n\textbackslash n[doc]}
    & Summarize the following text.\texttt{\textbackslash n\textbackslash n[doc]} \\
\midrule
Core
    & \selectlanguage{basque}Pentsatu ondo zein den hurrengo testuaren edukirik garrantzitsuena eta laburtu testua edukirik garrantzitsuena erabiliz.\texttt{\textbackslash n\textbackslash n[doc]}
    & \selectlanguage{spanish}Piensa bien cuál es el contenido más relevante en el siguiente texto y resúmelo usando el contenido más relevante.\texttt{\textbackslash n\textbackslash n[doc]}
    & Think well about what the most important content of the following text is and summarize the text using the most important content.\texttt{\textbackslash n\textbackslash n[doc]} \\
\midrule
5W1H
    & \selectlanguage{basque}Laburtu hurrengo testua, 5W1H metodoa erabiliz (zer, nork, noiz, non, zergatik, nola).\texttt{\textbackslash n\textbackslash n[doc]}
    & \selectlanguage{spanish}Resume el siguiente texto usando el método de las 5W1H (qué, quién, cuándo, dónde, por qué, cómo).\texttt{\textbackslash n\textbackslash n[doc]}
    & Summarize the following text using the 5W1H method (what, who, when, where, why and how).\texttt{\textbackslash n\textbackslash n[doc]} \\
\midrule
tldr
    & \texttt{[doc]\textbackslash n\textbackslash n}tldr: 
    & \texttt{[doc]\textbackslash n\textbackslash n}tldr:
    & \texttt{[doc]\textbackslash n\textbackslash n}tldr: \\
\bottomrule
\end{tblr}
    \caption{Prompts used to collect summaries from LLMs. Note that \texttt{[doc]} is a placeholder for the actual source document, i.e., the original newspaper article.}
    \label{tab:prompts}
\end{table*}

\begin{figure*}[htb]
    \centering
    \include{figures/prometheus}
    \vspace{-1em}
    \caption{Prompt from Prometheus-Eval used for LLM-as-a-Judge evaluation. Note that \texttt{instruction}, \texttt{response}, \texttt{reference answer}, and \texttt{rubric} are placeholders that are filled with specific information for each example.}
    \label{fig:prometheus}
\end{figure*}

\section{Annotation Guidelines}
\label{ann:guidelines}

\subsection*{Coherence}
Coherence refers to the collective quality of all sentences. We align this dimension with the DUC quality question \cite{Dang2005OverviewOD} of structure and coherence whereby ``the summary should be well-structured and well-organized. The summary should not just be a heap of related information, but should build from sentence to sentence to a coherent body of information about a topic''. To clarify, does the summary have a coherent flow of ideas and does it have connectors to explicitly define the relationships between ideas? It does not matter if there are grammatical errors or incorrect information. If the summary is simply a list of events, annotators should penalize it. Annotators do not penalize single sentences for not having connectors.

Coherence is rated 1-5 according to the following guidelines:
 \begin{enumerate}[noitemsep,label={\bfseries\arabic*}:]
     \item The summary contains a bullet point list of events and there is no internal consistency (e.g., random words).
     \item The summary contains a bullet point list of events, but there is no consistency between the phrases in each bullet point (e.g., 1. Keyboard, 2. Watching TV).
     \item The summary contains a bullet point list of events, but the phrases in each bullet point are well developed.
     \item The summary contains implicit coherence, but not explicit. That means that there are no temporal connectors or discourse markers, but the sentences have a temporal/causal flow.
     \item The summary contains explicit coherence via textual resources, such as temporal connectors or discourse markers that define the relationship between ideas and sentences.
 \end{enumerate}

\subsection*{Consistency}
Consistency refers to the factual alignment between the summary and the summarized source. A factually consistent summary contains only statements that are entailed by the source document. Annotators should also penalize summaries that contain hallucinated facts. If the summary contains information not found in the original document, annotators should penalize it. For temporal expressions (today, yesterday, this year), if the expression is consistent with the original information, annotators should assume that the summary is consistent and should not penalize.

Consistency is rated 1-5 according to the following guidelines:
 \begin{enumerate}[noitemsep,label={\bfseries\arabic*}:]
     \item The summary does not contain any ideas from the original text.
     \item The summary contains a large amount of incorrect information.
     \item The summary contains several incorrect pieces of information.
     \item The summary contains one incorrect piece of information.
     \item The summary is completely factual.
 \end{enumerate}

\subsection*{Fluency}
Fluency refers to the quality of individual sentences. Drawing again from the DUC quality guidelines, sentences in the summary ``should have no formatting problems, capitalization errors or obviously ungrammatical sentences (e.g., fragments, missing components) that make the text difficult to read''. Is the summary well written following the grammar and spelling conventions for the language? If it has grammatical or spelling errors, annotators should penalize it.  However, this criteria only takes grammar and fluency into account. While the style can sometimes be forced, if it is grammatically correct, we give full points.

Fluency is rated 1-5 according to the following guidelines:
\begin{enumerate}[noitemsep,label={\bfseries\arabic*}:]
    \item There are many important grammatical errors or the summary mixes language varieties/dialects in a very unnatural and uncomfortable way or the summary is not written in the language it is asked for.
    \item There are various important grammatical errors or the summary mixes language varieties/dialects in an unnatural and uncomfortable way.
    \item There are grammatical or spelling errors of lesser importance or the summary mixes language varieties/dialects in a somewhat unnatural and uncomfortable way.
    \item There are few grammatical or spelling errors.
    \item There is no grammatical or spelling errors and the variety of language/dialect used is coherent.
\end{enumerate}

\subsection*{Relevance}
Relevance refers to the selection of important content from the source. The summary should include only important information from the source document. Annotators should penalize summaries that contain redundancies and excess information. They should also penalize metalinguistic phrases generated by the model, such as `Here is the summary:' (-1). Finally, they should also penalize the presence of subjective opinions if they are not part of the original text (-1).

Relevance is rated 1-5 according to the following guidelines:
\begin{enumerate}[noitemsep,label={\bfseries\arabic*}:]
    \item The summary is the original text or only contains irrelevant information.
    \item The summary contains a large amount of irrelevant information.
    \item The summary contains a mix of relevant and irrelevant information.
    \item The summary contains mostly relevant information.
    \item All information in the summary is relevant.
\end{enumerate}

\subsection*{5W1H}
5W1H refers to the maintenance of all important information (the Ws in 5W1H---who, what, when, where, why, how) from the source document. The summary should not lack any important information available in the source.

5W1H is rated 1-5 according to the following guidelines:
\begin{enumerate}[noitemsep,label={\bfseries\arabic*}:]
    \item The summary contains no relevant W from the original text.
    \item The summary contains only 1-2 Ws.
    \item The summary lacks several relevant Ws.
    \item The summary is lacking one relevant W.
    \item The summary contains all the relevant Ws from the original text.
\end{enumerate}

\section{Dataset Information}
\label{ann:data_info}

\subsection{Annotation Examples}
\label{ann:examples}

Figures \ref{fig:eu-summ-ex} and \ref{fig:es-summ-ex} provide an example of low- and high-quality summaries in Spanish, along with their corresponding average scores across all five evaluation criteria.

\begin{figure*}[htb]
    \centering
    \begin{tcolorbox}[enhanced,title={Original news article},width=\textwidth,fonttitle=\sffamily\bfseries\scriptsize,fontupper=\small,fontlower=\small,left=1mm,right=1mm,halign=left,before upper=\parindent8pt]
    \selectlanguage{basque}
    \noindent Emakume batek sexu eraso bat jasan zuela salatu zuen atzo goizaldean, Etxebarrin (Bizkaia). Herriko jaiak ospatzen ari dira, eta kalean jarri ohi dituzten komunetan jazo zen erasoa. Etxebarriko Udalak bertan behera utzi zituen atzorako antolatuta zeuden ekitaldi guztiak. Udalak, sanantonioetako jai batzordeak eta txosnek elkarretaratzea egin dute Zintururi plazan, eta ehunka herritar bildu dira `Erasorik ez, erantzunik gabe' lelopean. Ertzaintzak ikerketa abiatu du, eta ez dute inor atxilotu. \par
    «Gure jaiak elkartzeko eta dibertitzeko guneak dira, eta gozatu nahi duten pertsona guztiak gonbidatzen ditugu, betiere askatasuna eta bizikidetza errespetatuz. Mezu hau ulertzen ez duten eta bestelako helburuak dituzten pertsonak ez dira ongi etorriak», adierazi du udalak ohar bidez.
\end{tcolorbox}%
\begin{tcbraster}[raster columns=2,raster equal height,nobeforeafter,raster column skip=3mm]
\begin{tcolorbox}[enhanced,title={Summary: Command R+ with Core prompt},fonttitle=\sffamily\bfseries\scriptsize,fontupper=\small,fontlower=\small,left=1mm,right=1mm,halign=left,before lower=\parindent8pt]
    \begin{tblr}{width=\linewidth,colspec={*{6}{X}},row{1}={c,font=\sffamily\bfseries\scriptsize},row{2}={c,font=\ttfamily\small},rowsep=0pt,colsep=0pt,column{6}={bg=red9}}
        Coh & Con & Flu & Rel & 5W1H & Avg \\
        2.00 & 2.67 & 1.67 & 2.00 & 2.00 & 2.07 \\
    \end{tblr}
    \tcblower
    \selectlanguage{basque}
    \noindent Testu honetan, garrantzitsuena da herritarren erantzuna eta errespetua eskatzea. Horregatik, laburtutako testuan, hurrengo frasiak eraginkorrak izan daitezke: \par
    ``Erasorik ez, erantzunik gabe'' lelopea herritarren eragile eta konprometituta dagoena erakusten du. Herriko jai honetan, emakumeen eta pertsonen segurtasuna babesteko, herritar guztiak erantzun bat eman behar dute. \par
    Udalak ere, adierazpen baliagarriak egin ditu, ulertu ez duten eta erasoak egiten dituzten pertsonak ez dira ongi etorriak. Horrela, herriko segurtasun eta askatasun boterea eraginkorra sustatzen du. \par
    Bertan behera utzi zituen ekitaldi guztiak, udalak erantzukizuna erakusten du, eta herriko jai honetan segurtasun eta askatasun inguruko normak betetzen dituela adierazi nahi du. \par
    Horregatik, garrantzitsuena da herritarren erantzuna eta errespetua eskatzea, eta udalak ere, erantzukizuna hartzen duen adierazpen baliagarriak egin du.
\end{tcolorbox}%
\begin{tcolorbox}[enhanced,title={Summary: GPT-4o with Base prompt},fonttitle=\sffamily\bfseries\scriptsize,fontupper=\small,fontlower=\small,left=1mm,right=1mm,halign=left,before lower=\parindent8pt]
    \begin{tblr}{width=\linewidth,colspec={*{6}{X}},row{1}={c,font=\sffamily\bfseries\scriptsize},row{2}={c,font=\ttfamily\small},rowsep=0pt,colsep=0pt,column{6}={bg=azure9}}
        Coh & Con & Flu & Rel & 5W1H & Avg \\
        4.67 & 4.67 & 4.67 & 4.00 & 5.00 & 4.60 \\
    \end{tblr}
    \tcblower
    \selectlanguage{basque}
    \noindent Hona hemen testua laburtuta: \par
    Etxebarrin (Bizkaia), herriko jaietan, emakume batek sexu eraso bat jasan zuen atzo goizaldean kalean jarritako komunetan. Ondorioz, Udalak egun horretarako aurreikusitako ekitaldi guztiak bertan behera utzi zituen. Zintururi plazan elkarretaratzea egin zuten Udalak, jai batzordeak eta txosnek, eta ehunka herritar bildu ziren ``Erasorik ez, erantzunik gabe'' lelopean. Ertzaintzak ikerketa abiatu du, baina ez da inor atxilotu oraindik. Udalak ohar batean adierazi du jaiak bizikidetza eta askatasuna errespetatuz ospatu behar direla, eta erasoak egiten dituzten pertsonak ez direla ongi etorriak.
\end{tcolorbox}
\end{tcbraster}
    \vspace{-1em}
    \caption{Examples of low- and high-quality summaries of a news article in Basque. Scores are the average of three annotators. Source of the article: Berria.\protect\footnotemark[15]}
    \label{fig:eu-summ-ex}
\end{figure*}
\setcounter{footnote}{15} 

\begin{figure*}[htb]
    \centering
    \begin{tcolorbox}[enhanced,title={Original news article},width=\textwidth,fonttitle=\sffamily\bfseries\scriptsize,fontupper=\small,fontlower=\small,left=1mm,right=1mm,halign=left,before upper=\parindent8pt]
    \selectlanguage{spanish}
    \noindent La Policía ha detenido en Alemania al joven de 17 años que estaba siendo buscado como presunto autor de la muerte de su hermana , de 14 , cuyo cadáver fue hallado en el domicilio familiar de Mataró , Barcelona , el pasado junio . El joven ha sido trasladado ya a España y ha sido puesto a disposición de la fiscalía de menores , según fuentes cercanas al caso . El menor estaba siendo buscado desde el pasado mes de junio , después de que se encontrara en el domicilio familiar el cadáver de su hermana , de 14 años , con heridas de arma blanca . El crimen ocurrió cuando la madre de ambos menores , de origen ruso , se encontraba de viaje , por lo que los dos adolescentes se habían quedado solos en la vivienda , situada en el barrio Cerdanyola de Mataró . Alertada por no poder contactar con sus hijos , la mujer pidió a unos amigos que acudieran a su domicilio , donde estos encontraron el cadáver de la menor , sin rastro de su hermano mayor.
\end{tcolorbox}%
\begin{tcbraster}[raster columns=2,raster equal height,nobeforeafter,raster column skip=3mm]
\begin{tcolorbox}[enhanced,title={Summary: Llama 3.1 70B Instruct with 5W1H prompt},fonttitle=\sffamily\bfseries\scriptsize,fontupper=\small,fontlower=\small,left=1mm,right=1mm,halign=left,before lower=\parindent8pt]
    \begin{tblr}{width=\linewidth,colspec={*{6}{X}},row{1}={c,font=\sffamily\bfseries\scriptsize},row{2}={c,font=\ttfamily\small},rowsep=0pt,colsep=0pt,column{6}={bg=red9}}
        Coh & Con & Flu & Rel & 5W1H & Avg \\
        3.00 & 3.67 & 5.00 & 3.67 & 4.67 & 4.00 \\
    \end{tblr}
    \tcblower
    \selectlanguage{spanish}
    \noindent Aquí está el resumen del texto usando el método de las 5W1H: \par
    Qué: La Policía ha detenido al joven de 17 años como presunto autor de la muerte de su hermana de 14 años, cuyo cadáver fue hallado en su domicilio familiar en Mataró, Barcelona, con heridas de arma blanca. \par
    Quién: El joven de 17 años, cuya identidad no ha sido revelada debido a su condición de menor de edad. \par
    Cuándo: El crimen ocurrió en junio pasado, y el joven fue detenido recientemente y trasladado a España para ser puesto a disposición de la fiscalía de menores. \par
    Dónde: En el domicilio familiar situado en el barrio Cerdanyola de Mataró, Barcelona. \par
    Por qué: Las razones detrás del crimen aún no están claras y son objeto de investigación. \par
    Cómo: El joven supuestamente atacó a su hermana con un arma blanca, causándole heridas que resultaron fatales. \par
    Este resumen proporciona una visión general concisa del incidente utilizando las preguntas básicas de las 5W1H para transmitir la información clave.
\end{tcolorbox}%
\begin{tcolorbox}[enhanced,title={Summary: GPT-4o with Base prompt},fonttitle=\sffamily\bfseries\scriptsize,fontupper=\small,fontlower=\small,left=1mm,right=1mm,halign=left,before lower=\parindent8pt]
    \begin{tblr}{width=\linewidth,colspec={*{6}{X}},row{1}={c,font=\sffamily\bfseries\scriptsize},row{2}={c,font=\ttfamily\small},rowsep=0pt,colsep=0pt,column{6}={bg=azure9}}
        Coh & Con & Flu & Rel & 5W1H & Avg \\
        4.33 & 5.00 & 5.00 & 4.67 & 5.00 & 4.80 \\
    \end{tblr}
    \tcblower
    \selectlanguage{spanish}
    \noindent La Policía ha detenido en Alemania a un joven de 17 años, sospechoso de haber matado a su hermana de 14 años en Mataró, Barcelona, el pasado junio. El joven ya ha sido trasladado a España y puesto a disposición de la fiscalía de menores. El crimen ocurrió cuando su madre estaba de viaje y los adolescentes se quedaron solos en su casa. La madre, al no poder contactar con ellos, pidió a unos amigos que fueran a su domicilio, donde encontraron el cadáver de la menor con heridas de arma blanca y sin rastro del hermano.
\end{tcolorbox}
\end{tcbraster}
    \vspace{-1em}
    \caption{Examples of low- and high-quality summaries of a news article in Spanish. Scores are averages of three annotators. Source of the article: El País.\protect\footnotemark[16]}
    \label{fig:es-summ-ex}
\end{figure*}

\subsection{Inter-annotator Agreement}
\label{ann:iia}

\begin{figure*}[htb]
    \centering
    \includegraphics[width=.8\textwidth,trim={-2pt 6pt 0 -8pt},clip]{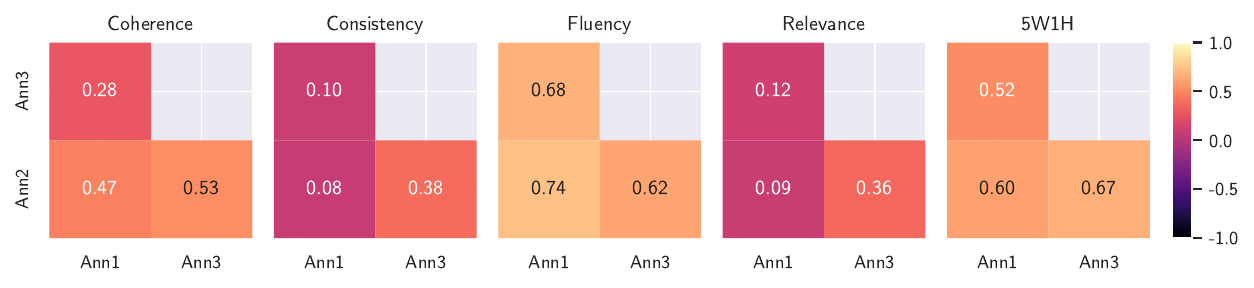}
    \caption{Quadratic Cohen's $\kappa$ in Basque R\#3}
    \label{fig:eu-round3-kappa}
    \par
    \includegraphics[width=.8\textwidth,trim={-2pt 6pt 0 -8pt},clip]{images/kappa-es-round3.pdf}
    \caption{Quadratic Cohen's $\kappa$ in Spanish R\#3}
    \label{fig:es-round3-kappa}
    \par
    \includegraphics[width=.8\textwidth,trim={-2pt 6pt 0 -8pt},clip]{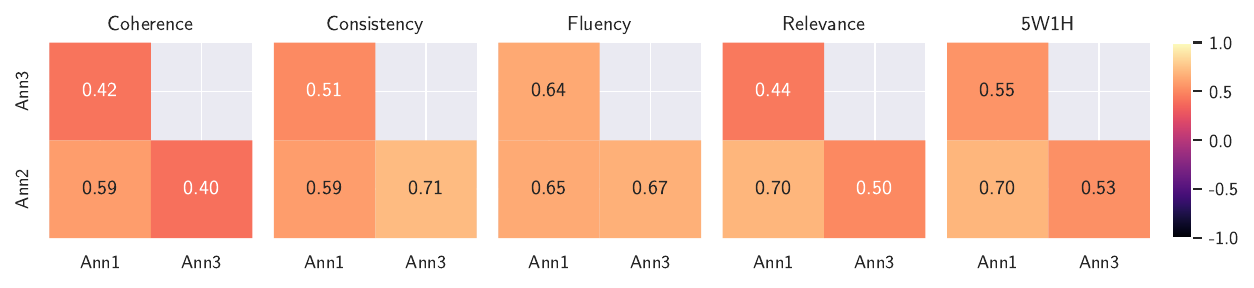}
    \caption{Agreement percentage in Basque R\#3}
    \label{fig:eu-round3-perc}
    \par
    \includegraphics[width=.8\textwidth,trim={-2pt 6pt 0 -8pt},clip]{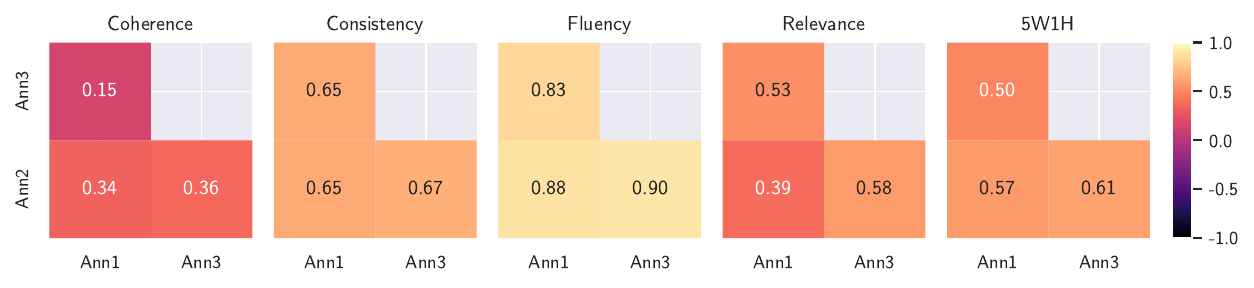}
    \caption{Agreement percentage in Spanish R\#3 }
    \label{fig:es-round3-perc}
    \par
    \includegraphics[width=.8\textwidth,trim={5pt 6pt 0 -8pt},clip]{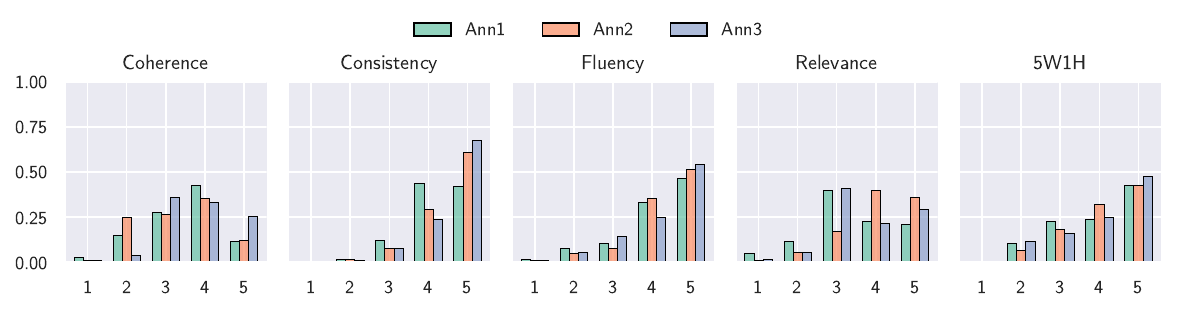}
    \caption{Score distribution across annotators in Basque R\#3 }
    \label{fig:eu-round3-dist}
    \par
    \includegraphics[width=.8\textwidth,trim={5pt 6pt 0 -8pt},clip]{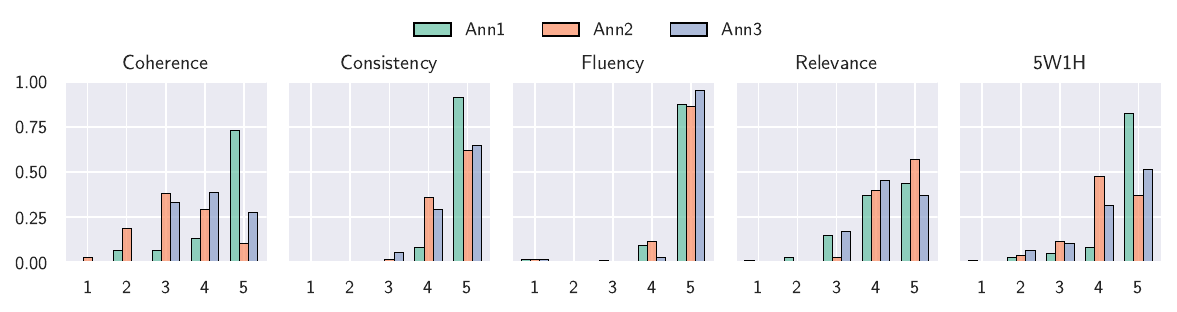}
    \caption{Score distribution across in Spanish R\#3 }
    \label{fig:es-round3-dist}
\end{figure*}

Krippendorff's $\alpha$ correlations for each round and language are shown in Table \ref{tab:alphas}, while Figures \ref{fig:eu-round3-kappa} to \ref{fig:es-round3-dist} show inter-annotator agreements in terms of quadratic Cohen's $\kappa$, agreement percentage and score distribution for the final annotation round (Round \#3) in Basque and Spanish.


\subsection{Dataset Statistics}
\label{ann:basse-stats}

Table \ref{tab:basse-detail} shows detailed statistics of BASSE, with summary statistics broken down by author model and prompt. Figure \ref{fig:data-stats} shows the distribution of compression ratios and novel 2-grams for different summarization approaches. Compression ratio indicates summary conciseness with respect to the source article \cite{segarra-soriano-etal-2022-dacsa}, while novel 2-grams measure the percentage of new word pairs in the summary \cite{kryscinski-etal-2018-improving}.

\begin{table*}[htb]
    \centering
    \begin{tblr}{
       colspec={*{11}{r}},
       cells={font=\small},
       columns={rightsep=5pt},
       row{1,2}={c,m,font=\small\bfseries},
       column{1}={l,font=\small\bfseries},
       column{1,6}={rightsep=10pt},
       rowsep=0pt
    }
    \toprule
                  & \SetCell[c=5]{c} Basque & & & & & \SetCell[c=5]{c} Spanish \\
                   \cmidrule[lr]{2-6} \cmidrule[lr]{7-11}
                  & Doc & ¬ eu & Tok & Voc & {s/d}           &  Doc & ¬ es &  Tok & Voc & {s/d} \\
    \midrule
        Source article        &  45 &   - & 22,862 & 7,328 & 34 &  45 &   - & 38,551 & 8,200 & 30 \\
    \midrule
        Subhead               &  45 &   0 &   1,562 &    927 &  3 &  45 &   0 &   1,040 &    578 & 1 \\
        Human summary         &  75 &   0 &   8,867 &  3,428 &  6 &  75 &   0 &  11,635 &  3,180 & 5 \\
        LLM summary           & 900 &  54 & 133,755 & 15,755 & 10 & 900 &  27 & 158,945 & 10,844 & 8 \\
        \quad \textit{\scriptsize by model} \\
        \quad \quad Claude    & 180 &  38 & 29,602 & 7,589 & 12 & 180 &  26 & 32,301 & 6,226 & 10 \\
        \quad \quad Cmd R+    & 180 &  16 & 37,236 & 7,945 & 13 & 180 &   1 & 30,460 & 4,724 &  7 \\
        \quad \quad GPT-4o    & 180 &   0 & 30,859 & 7,235 & 12 & 180 &   0 & 29,647 & 4,962 &  7 \\
        \quad \quad Llama     & 180 &   0 & 14,671 & 3,279 &  6 & 180 &   0 & 33,918 & 4,401 &  8 \\
        \quad \quad Reka      & 180 &   0 & 21,387 & 4,658 &  7 & 180 &   0 & 32,619 & 4,858 &  7 \\
        \quad \textit{\scriptsize by prompt} \\
        \quad \quad Base      & 225 &  22 & 43,452 & 9,965 & 12 & 225 &   0 & 42,242 & 5,836 &  8 \\
        \quad \quad Core      & 225 &   0 & 24,746 & 5,378 &  7 & 225 &   0 & 36,059 & 5,349 &  7 \\
        \quad \quad 5W1H      & 225 &   0 & 37,273 & 6,740 & 15 & 225 &   0 & 49,885 & 5,625 & 10 \\
        \quad \quad tldr      & 225 &  32 & 28,284 & 7,483 &  7 & 225 &  27 & 30,759 & 6,099 &  6 \\
    \bottomrule
\end{tblr}%
    \caption{Statistics of the BASSE corpus, including the source articles and their summaries. We report the number of documents (Doc)---and, of those, how many are not in the target language (¬ eu or ¬ es)---, total tokens (Tok), vocabulary size (Voc), and sentences per document (s/d).}
    \label{tab:basse-detail}
\end{table*}


\subsection{Summarization Evaluation}
\label{ann:summarization-results}

Table \ref{tab:human-eval-detail} presents the average human evaluation scores for summaries on a 5-point Likert scale across the five evaluation criteria: \textit{coherence} (Coh), \textit{consistency} (Con), \textit{fluency} (Flu), \textit{relevance} (Rel), and \textit{5W1H} coverage. The table reports scores for the original article subheads, human-written summaries, and the summaries generated by five different LLMs with four prompting strategies.

\begin{table*}[htb]
    \centering
    \begin{tblr}{
       width=\textwidth,
       colspec={*{13}{c}},
       cells={font=\small},
       columns={rightsep=3pt},
       row{1,2}={c,font=\small\bfseries},
       column{1}={l,m,font=\small\bfseries},
       column{1,7}={rightsep=6pt},
       rowsep=0pt,
   }
   \toprule
        &  \SetCell[c=6]{c} Basque & & & & & & \SetCell[c=6]{c} Spanish \\
    \cmidrule[lr]{2-7} \cmidrule[lr]{8-13}
        &  Coh & Con & Flu & Rel & 5w1h & Avg & Coh & Con & Flu & Rel & 5w1h & Avg \\
   \midrule
    Subhead & 3.70 & 4.79 & \pop{4.97} & \pop{4.56} & 2.80 & 4.16 & 4.60 & 4.85 & \pop{4.93} & 4.32 & 2.01 & 4.14 \\
    Human              & \pop{4.77} & \pop{4.94} & 4.85 & 4.51 & \pop{4.69} & \pop{4.75} & \pop{4.91} & \pop{4.88} & 4.80 & \pop{4.48} & \pop{4.34} & \pop{4.68} \\
    LLM                & 3.46 & 4.29 & 3.87 & 3.74 & 4.06 & 3.89 & 3.88 & 4.68 & 4.82 & 4.14 & 4.18 & 4.34 \\
   \quad \textit{\scriptsize by model} \\ 
   \quad \quad Claude    & 3.17 & \pop{4.61} & 3.83 & 3.54 & 4.57 & 3.95 & 3.17 & 4.71 & 4.40 & 3.81 & 4.40 & 4.10 \\
   \quad \quad GPT-4o    & \pop{3.88} & 4.52 & \pop{4.55} & 3.78 & \pop{4.62} & \pop{4.27} & 4.13 & 4.77 & \pop{4.94} & \pop{4.37} & \pop{4.37} & \pop{4.52} \\
   \quad \quad Reka      & 3.72 & 4.20 & 4.05 & 4.02 & 4.14 & 4.03 & \pop{4.21} & 4.49 & 4.87 & 4.26 & 4.17 & 4.40 \\
   \quad \quad LLama 3.1 & 3.59 & 4.50 & 4.29 & \pop{4.31} & 3.45 & 4.03 & 3.78 & \pop{4.79} & 4.93 & 3.98 & 3.94 & 4.29 \\
   \quad \quad Cmd R+    & 2.95 & 3.65 & 2.64 & 3.07 & 3.53 & 3.17 & 4.10 & 4.66 & \pop{4.94} & 4.29 & 4.02 & 4.40 \\
   \quad \textit{\scriptsize by prompt} \\
   \quad \quad Base      & 3.78 & 4.20 & 3.70 & 3.36 & 4.19 & 3.85 & \pop{4.22} & 4.67 & 4.91 & 4.12 & 4.20 & \pop{4.42} \\
   \quad \quad Core       & 3.61 & 4.33 & 4.05 & 3.73 & 3.64 & 3.87 & 4.11 & 4.73 & 4.93 & 4.08 & 3.94 & 4.36 \\
   \quad \quad 5W1H      & 2.62 & 4.22 & \pop{4.10} & 3.64 & \pop{4.39} & 3.79 & 3.03 & 4.58 & \pop{4.96} & 4.03 & \pop{4.65} & 4.25 \\
   \quad \quad tldr      & \pop{3.82} & \pop{4.42} & 3.64 & \pop{4.24} & 4.01 & \pop{4.03} & 4.14 & \pop{4.75} & 4.47 & \pop{4.35} & 3.94 & 4.33 \\
   \bottomrule
\end{tblr}%
    \caption{Mean human evaluation scores on a 5-point Likert scale for summaries across the five evaluation criteria, and their absolute average (Avg). Highest scores for each criterion and computing class are highlighted in \pop{bold}.}
    \label{tab:human-eval-average}
\end{table*}

\begin{table*}[htb]
    \centering
    \begin{tblr}{
       width=\textwidth,
       colspec={*{13}{c}},
       cells={font=\small},
       columns={rightsep=5pt},
       row{1,2}={c,font=\small\bfseries},
       column{1}={l,m,font=\small\bfseries},
       column{1,7}={rightsep=10pt},
       rowsep=0pt,
   }
   \toprule
      & \SetCell[c=6]{c} Basque & & & & & & \SetCell[c=6]{c} Spanish \\
    \cmidrule[lr]{2-7} \cmidrule[lr]{8-13}
      & Coh & Con & Flu & Rel & 5w1h & Avg & Coh & Con & Flu & Rel & 5w1h & Avg \\
   \midrule
   Claude \\
   \quad Base & 3.20 & 4.71 & 3.13 & 3.04 & 4.60 & 3.74 & 3.43 & 4.73 & 4.96 & 3.86 & 4.34 & 4.26 \\
   \quad Core  & 3.73 & 4.57 & 4.47 & 3.76 & 4.30 & 4.17 & 3.59 & 4.81 & 4.96 & 3.81 & 4.21 & 4.28 \\
   \quad 5W1H & 2.51 & 4.36 & \pop{4.61} & 3.45 & \pop{4.85} & 3.96 & 2.67 & 4.59 & 4.98 & 3.80 & \pop{4.76} & 4.16 \\
   \quad tldr & 3.24 & \pop{4.80} & 3.10 & 3.92 & 4.53 & 3.92 & 3.00 & 4.70 & 2.69 & 3.78 & 4.30 & 3.69 \\
   GPT-4o \\
   \quad Base & \pop{4.41} & 4.57 & 4.59 & 2.61 & 4.83 & 4.20 & 4.53 & 4.78 & 4.96 & 4.34 & 4.33 & 4.59 \\
   \quad Core  & 4.19 & 4.50 & 4.56 & 4.36 & 4.27 & 4.37 & 4.48 & 4.80 & 4.93 & 4.39 & 4.30 & 4.58 \\
   \quad 5W1H & 2.82 & 4.41 & 4.56 & 3.56 & 4.81 & 4.03 & 3.05 & 4.71 & 4.93 & 4.18 & 4.65 & 4.31 \\
   \quad tldr & 4.08 & 4.59 & 4.47 & 4.60 & 4.58 & \pop{4.46} & 4.47 & 4.79 & 4.94 & \pop{4.57} & 4.21 & \pop{4.60} \\
   Reka \\
   \quad Base & 4.12 & 4.10 & 3.96 & 3.91 & 4.40 & 4.10 & \pop{4.64} & 4.39 & 4.83 & 4.26 & 4.16 & 4.46 \\
   \quad Core  & 3.88 & 4.10 & 4.12 & 3.73 & 3.85 & 3.94 & 4.48 & 4.63 & 4.84 & 4.13 & 4.03 & 4.42 \\
   \quad 5W1H & 2.81 & 4.19 & 4.13 & 3.87 & 4.32 & 3.86 & 3.30 & 4.33 & 4.93 & 4.10 & 4.63 & 4.25 \\
   \quad tldr & 4.07 & 4.39 & 3.99 & 4.57 & 3.97 & 4.20 & 4.41 & 4.61 & 4.87 & 4.54 & 3.87 & 4.46 \\
   Llama 3.1 \\
   \quad Base & 3.94 & 4.33 & 4.31 & 4.47 & 3.39 & 4.09 & 4.07 & 4.82 & 4.87 & 3.95 & 4.09 & 4.36 \\
   \quad Core  & 3.67 & 4.61 & 4.27 & 4.02 & 3.01 & 3.92 & 3.53 & 4.79 & 4.93 & 3.61 & 3.73 & 4.12 \\
   \quad 5W1H & 2.64 & 4.49 & 4.23 & 3.83 & 4.09 & 3.86 & 3.09 & 4.75 & 4.95 & 3.87 & 4.54 & 4.24 \\
   \quad tldr & 4.09 & 4.56 & 4.36 & \pop{4.90} & 3.30 & 4.24 & 4.42 & 4.81 & 4.98 & 4.50 & 3.41 & 4.43 \\
   Cmd R+ \\
   \quad Base & 3.25 & 3.31 & 2.50 & 2.79 & 3.75 & 3.12 & 4.44 & 4.61 & 4.92 & 4.17 & 4.07 & 4.44 \\
   \quad Core  & 2.56 & 3.88 & 2.82 & 2.78 & 2.79 & 2.97 & 4.49 & 4.64 & 4.97 & 4.44 & 3.45 & 4.40 \\
   \quad 5W1H & 2.33 & 3.64 & 2.95 & 3.50 & 3.88 & 3.26 & 3.06 & 4.53 & \pop{5.00} & 4.18 & 4.67 & 4.29 \\
   \quad tldr & 3.64 & 3.78 & 2.30 & 3.21 & 3.69 & 3.32 & 4.42 & \pop{4.84} & 4.86 & 4.36 & 3.90 & 4.48 \\
   \bottomrule
\end{tblr}%
    \caption{Mean human evaluation scores on a 5-point Likert scale for summaries across the five evaluation criteria, and their absolute average (Avg). Highest scores for each criterion and computing class are highlighted in \pop{bold}.}
    \label{tab:human-eval-detail}
\end{table*}

\section{Metrics}

\subsection{Metric information}
\label{ann:metrics}

\begin{itemize}[noitemsep,wide,topsep=0pt]
    \item \textbf{ROUGE} \cite{lin-2004-rouge} is a series of metrics that calculate token subsequence overlap between a candidate summary and a set of reference summaries. We calculate ROUGE-\{1--4\}, ROUGE-L, and ROUGE-su* variants.
    \item \textbf{ROUGE-we} \cite{ng-abrecht-2015-better} extends the original ROUGE metric by incorporating soft matching based on cosine similarity of word embeddings. We use FastText embeddings for both Basque and Spanish\footnote{\url{https://fasttext.cc/docs/en/crawl-vectors}}.
    \item \textbf{BertScore} \cite{Zhang2019BERTScoreET} calculates the similarity of two texts as the average token-level cosine similarities between greedily aligned tokens in each text. For our experiments, we replace BERT with multilingual BERT base uncased\footnote{\url{https://huggingface.co/google-bert/bert-base-multilingual-uncased}} and report precision (p), recall (r), and F1 (f) scores.
    \item \textbf{BLEU} \cite{papineni-etal-2002-bleu} is a corpus-level metric originally designed for machine translation evaluation. It calculates average n-gram precision, coupled with a brevity penalty. We specifically use the SacreBLEU implementation \cite{post-2018-call}.
    \item \textbf{CHRF} \cite{popovic-2017-chrf} calculates character-level n-gram overlaps between the candidate summary and the reference summary.
    \item \textbf{METEOR} \cite{banerjee-lavie-2005-meteor} computes an alignment between candidate and reference sentences, mapping unigrams in the generated summary to 0 or 1 unigrams in the reference, based on stems, synonyms, and paraphrase matches. Precision and recall are computed and reported as a harmonic mean. We compute Multilingual METEOR (mMETEOR) by setting the language to \texttt{other} and only lower-casing the input.
    \item \textbf{CIDEr} \cite{vedantam2015-cider} is a corpus-level metric originally designed for image description evaluation. It calculates the average cosine similarity between TD-IDF weighted n-gram representations of the candidate summary and a set of reference summaries, taking the average score of 1--4 grams. As this metric applies stemming, we use the Basque and Spanish versions of the Snowball Stemmer \cite{snowball}.
    \footnotetext[15]{\url{https://www.berria.eus/euskal-herria/ehunka-herritarrek-etxebarriko-sexu-erasoa-salatu-dute_2126343_102.html}}
    \item \textbf{Data Statistics} \cite{grusky-etal-2018-newsroom} calculates a series of metrics originally designed to describe how extractive a dataset is. This includes \textit{i)} Coverage, which measures how derivative a summary is; \textit{ii)} Density, which measures how well a summary can be described as a series of extractions from the original text or \textit{iii)} Compression, which is the ratio of the length of the summary divided by the length of the original text. SummEval further introduces n-gram related novelty---novel \{1--3\} gram---and redundancy scores---repeated \{1--3\} gram. Finally, we also include the length of the generated summaries.
\end{itemize}

\footnotetext[16]{\url{http://elpais.com/ccaa/2019/09/28/catalunya/1569665675\_576416.html}}

\subsection{Metric Correlation}
\label{ann:correlation-evaluation}

Tables \ref{tab:correlation-eu} and \ref{tab:correlation-es} present correlation values between automatic evaluation metrics and human annotations for Basque and Spanish respectively. For each automatic metric and human annotation criterion, both Kendall's $\tau$ and Spearman's $\rho$ rank correlation coefficients are reported. Rank correlation coefficients have been computed at model level.

\begin{table*}[htb]
    \centering
    \begin{tblr}{
       width=\textwidth,
       colspec={*{11}{r}},
       cells={font=\scriptsize},
       columns={rightsep=4pt},
       row{1,2}={c,font=\small\bfseries},
       column{1}={l,m,font=\small\bfseries},
       column{1,6}={rightsep=8pt},
       rowsep=0pt,
    }
    \toprule
            & \SetCell[c=5]{c} Kendall's $\tau$ & & & & & \SetCell[c=5]{c} Spearman's $\rho$ \\
            \cmidrule[lr]{2-6} \cmidrule[lr]{7-11}
            & Coh & Con & Flu & Rel & 5W1H & Coh & Con & Flu & Rel & 5W1H \\
    \midrule
    ROUGE-1      & 0.432 & 0.111 & 0.354 & 0.358 & 0.242 & 0.617 & 0.082 & 0.556 & 0.459 & 0.290 \\
    ROUGE-2      & 0.453 & 0.216 & \pop{0.501} & 0.484 & 0.095 & 0.630 & 0.223 & \pop{0.677} & 0.534 & 0.069 \\
    ROUGE-3      & 0.332 & 0.201 & \pop{0.455} & 0.364 & 0.005 & 0.463 & 0.194 & 0.606 & 0.474 & -0.042 \\
    ROUGE-4      & 0.295 & 0.164 & 0.364 & 0.305 & -0.105 & 0.409 & 0.148 & 0.487 & 0.421 & -0.179 \\
    ROUGE-L      & 0.537 & 0.111 & 0.311 & 0.526 & 0.032 & 0.738 & 0.126 & 0.496 & 0.683 & 0.015 \\
    ROUGE-su*    & 0.505 & 0.121 & 0.343 & 0.453 & 0.232 & 0.695 & 0.181 & 0.557 & 0.552 & 0.271 \\
    mROUGE-we    & 0.442 & 0.111 & 0.290 & 0.411 & 0.000 & 0.689 & 0.088 & 0.451 & 0.523 & -0.027 \\
    \midrule
    BertScore-p  & 0.516 & 0.037 & 0.164 & \pop{0.568} & -0.263 & 0.746 & 0.065 & 0.243 & 0.737 & -0.395 \\
    BertScore-r  & 0.316 & 0.016 & 0.343 & 0.053 & 0.421 & 0.441 & -0.071 & 0.540 & 0.045 & 0.525 \\
    BertScore-f  & \pop{0.611} & 0.079 & 0.322 & 0.432 & 0.042 & \pop{0.814} & 0.097 & 0.475 & 0.543 & 0.033 \\
    \midrule
    BLEU         & 0.179 & 0.142 & 0.206 & 0.463 & -0.158 & 0.245 & 0.198 & 0.223 & 0.611 & -0.217 \\
    CHRF         & -0.147 & 0.132 & 0.090 & -0.011 & 0.400 & -0.269 & 0.190 & 0.108 & -0.032 & 0.523 \\
    mMETEOR      & 0.347 & 0.026 & 0.332 & 0.211 & 0.368 & 0.462 & -0.005 & 0.527 & 0.208 & 0.406 \\
    CIDEr        & 0.211 & 0.047 & 0.195 & 0.495 & -0.084 & 0.302 & 0.079 & 0.297 & 0.677 & -0.108 \\
    \midrule
    Length       & -0.274 & -0.259 & -0.216 & \pop{-0.684} & 0.337 & -0.408 & -0.317 & -0.229 & \pop{-0.887} & 0.420 \\
    Novel 1-gram & -0.326 & -0.142 & -0.206 & \pop{-0.611} & 0.305 & -0.537 & -0.117 & -0.351 & \pop{-0.780} & 0.412 \\
    Novel 2-gram & -0.368 & -0.142 & -0.206 & \pop{-0.568} & 0.305 & -0.591 & -0.127 & -0.329 & \pop{-0.756} & 0.438 \\
    Novel 3-gram & -0.421 & -0.121 & -0.195 & -0.474 & 0.295 & -0.630 & -0.104 & -0.297 & -0.668 & 0.462 \\
    Rep. 1-gram  & -0.474 & -0.280 & -0.332 & -0.421 & -0.242 & -0.684 & -0.416 & -0.497 & -0.600 & -0.349 \\
    Rep. 2-gram  & -0.411 & \pop{-0.343} & -0.185 & -0.463 & -0.179 & -0.597 & \pop{-0.455} & -0.282 & -0.614 & -0.280 \\
    Rep. 3-gram  & -0.358 & -0.248 & -0.047 & -0.263 &   -0.168 & -0.519 & -0.370 & -0.128 & -0.451 & -0.299 \\
    \midrule
    Coverage     & 0.411 & 0.037 & 0.121 & 0.463 & -0.368 & 0.677 & 0.029 & 0.209 & 0.651 & -0.514 \\
    Compression  & 0.358 & 0.069 & 0.026 & 0.516 & -0.484 & 0.492 & 0.080 & -0.002 & 0.728 & -0.665 \\
    Density      & 0.379 & 0.121 & 0.195 & 0.432 & -0.295 & 0.603 & 0.125 & 0.300 & 0.597 & -0.474 \\
    \midrule
    Prom 2 7B & 0.106 & 0.287 & 0.047 & -0.016 & 0.417 & 0.171 & 0.421 & 0.091 & -0.045 & 0.551 \\
    Prom 2 8x7b & -0.070 & 0.272 & 0.179 & 0.101 & 0.600 & -0.112 & 0.366 & 0.248 & 0.077 & 0.767 \\
    M-Prom 7B & 0.132 & 0.360 & 0.240 & 0.176 & 0.656 & 0.186 & 0.492 & 0.368 & 0.228 & 0.810 \\
    M-Prom 14B & 0.263 & 0.269 & \pop{0.455} & 0.116 & \pop{0.695} & 0.344 & 0.389 & \pop{0.609} & 0.198 & \pop{0.829} \\
    Selene Mini & 0.187 & 0.223 & 0.049 & -0.112 & 0.684 & 0.231 & 0.291 & 0.088 & -0.148 & 0.812 \\
    Qwen2.5 7B & 0.080 & 0.122 & 0.086 & -0.257 & 0.488 & 0.149 & 0.170 & 0.173 & -0.404 & 0.654 \\
    Qwen2.5 72B & 0.112 & 0.032 & 0.106 & -0.229 & 0.526 & 0.196 & 0.064 & 0.183 & -0.375 & 0.676 \\
    GPT-4o mini  & \pop{0.660} & \pop{0.387} & 0.319 & -0.081 & \pop{0.744} & \pop{0.790} & \pop{0.509} & 0.467 & -0.097 & \pop{0.874} \\
    GPT-4o       & \pop{0.786} & \pop{0.430} & \pop{0.590} & 0.381 & \pop{0.702} & \pop{0.909} & \pop{0.574} & \pop{0.761} & 0.512 & \pop{0.859} \\
    \bottomrule
\end{tblr}
    \caption{Rank correlations between automatic metrics and human annotations for the Basque data. Largest three correlations for each criteria in \pop{bold}.}
    \label{tab:correlation-eu}
\end{table*}

\begin{table*}[htb]
    \centering
    \begin{tblr}{
       width=\textwidth,
       colspec={*{11}{r}},
       cells={font=\scriptsize},
       columns={rightsep=4pt},
       row{1,2}={c,font=\small\bfseries},
       column{1}={l,m,font=\small\bfseries},
       column{1,6}={rightsep=8pt},
       rowsep=0pt,
    }
    \toprule
            & \SetCell[c=5]{c} Kendall's $\tau$ & & & & & \SetCell[c=5]{c} Spearman's $\rho$ \\
            \cmidrule[lr]{2-6} \cmidrule[lr]{7-11}
            & Coh & Con & Flu & Rel & 5W1H & Coh & Con & Flu & Rel & 5W1H \\
    \midrule
    ROUGE-1      & 0.385 & 0.032 & -0.214 & 0.164 & 0.016 & 0.528 & 0.063 & -0.280 & 0.232 & 0.011 \\
    ROUGE-2      & 0.164 & 0.253 & -0.064 & 0.037 & -0.079 & 0.245 & 0.435 & -0.071 & 0.020 & -0.136 \\
    ROUGE-3      & 0.069 & 0.305 & 0.096 & -0.016 & -0.111 & 0.096 & \pop{0.478} & 0.102 & -0.003 & -0.232 \\
    ROUGE-4      & -0.005 & 0.284 & 0.139 & -0.016 & -0.079 & 0.028 & 0.435 & 0.206 & -0.055 & -0.186 \\
    ROUGE-L      & 0.491 & 0.263 & -0.257 & 0.364 & -0.322 & 0.675 & 0.394 & \pop{-0.343} & 0.475 & -0.479 \\
    ROUGE-su*    & 0.385 & 0.095 & -0.214 & 0.111 & -0.026 & 0.502 & 0.129 & -0.290 & 0.179 & -0.027 \\
    mROUGE-we    & 0.544 & 0.084 & \pop{-0.246} & 0.290 & -0.269 & \pop{0.752} & 0.102 & -0.310 & 0.460 & -0.426 \\
    \midrule
    BertScore-p  & \pop{0.596} & 0.232 & -0.021 & \pop{0.628} & -0.544 & \pop{0.774} & 0.322 & -0.051 & \pop{0.801} & -0.736 \\
    BertScore-r  & 0.449 & 0.116 & \pop{-0.310} & 0.142 & -0.047 & 0.572 & 0.168 & \pop{-0.369} & 0.223 & -0.091 \\
    BertScore-f  & 0.501 & 0.211 & -0.139 & \pop{0.480} & -0.396 & 0.699 & 0.307 & -0.225 & 0.617 & -0.582 \\
    \midrule
    BLEU         & 0.005 & 0.042 & 0.225 & 0.174 & -0.185 & 0.073 & 0.030 & 0.335 & 0.219 & -0.311 \\
    CHRF         & -0.121 & -0.200 & 0.064 & -0.090 & 0.406 & -0.157 & -0.233 & 0.131 & -0.105 & 0.543 \\
    mMETEOR      & 0.026 & 0.137 & -0.096 & -0.142 & 0.100 & 0.026 & 0.198 & -0.082 & -0.186 & 0.083 \\
    CIDEr        & \pop{0.565} & 0.084 & \pop{-0.300} & 0.459 & -0.417 & \pop{0.786} & 0.114 & \pop{-0.429} & \pop{0.654} & -0.593 \\
    \midrule
    Length       & -0.364 & -0.253 & 0.000 & -0.438 & 0.480 & -0.575 & -0.346 & 0.020 & -0.621 & 0.659 \\
    Novel 1-gram & -0.459 & -0.326 & -0.043 & \pop{-0.533} & 0.660 & -0.654 & -0.418 & -0.024 & \pop{-0.686} & 0.816 \\
    Novel 2-gram & -0.332 & \pop{-0.411} & 0.011 & -0.427 & 0.617 & -0.550 & \pop{-0.553} & 0.021 & -0.484 & 0.785 \\
    Novel 3-gram & -0.354 & \pop{-0.453} & -0.053 & -0.364 & 0.575 & -0.482 & \pop{-0.617} & -0.109 & -0.437 & 0.737 \\
    Rep. 1-gram  & -0.396 & -0.126 & 0.160 & -0.364 & 0.100 & -0.561 & -0.105 & 0.145 & -0.508 & 0.206 \\
    Rep. 2-gram  & -0.311 & -0.126 & 0.118 & -0.343 & 0.142 & -0.478 & -0.140 & 0.111 & -0.446 & 0.229 \\
    Rep. 3-gram  & -0.311 & -0.147 & 0.096 & -0.343 & 0.058 & -0.487 & -0.149 & 0.072 & -0.457 & 0.162 \\
    \midrule
    Coverage     & 0.480 & 0.242 & -0.021 & 0.470 & \pop{-0.702} & 0.659 & 0.317 & -0.026 & 0.618 & \pop{-0.848} \\
    Compression  & 0.322 & 0.316 & 0.011 & 0.332 & -0.586 & 0.539 & 0.445 & 0.009 & 0.466 & -0.792 \\
    Density      & 0.259 & \pop{0.337} & 0.107 & 0.332 & -0.522 & 0.388 & 0.474 & 0.131 & 0.397 & -0.682 \\
    \midrule
    Prom 2 7B & 0.037 & 0.005 & -0.139 & -0.159 & 0.328 & 0.027 & -0.007 & -0.170 & -0.251 & 0.403 \\
    Prom 2 8x7b & -0.215 & -0.128 & -0.033 & -0.064 & 0.639 & -0.354 & -0.194 & -0.048 & -0.096 & 0.826 \\
    M-Prom 7B & 0.148 & 0.026 & 0.086 & -0.107 & 0.525 & 0.091 & 0.109 & 0.139 & -0.140 & 0.598 \\
    M-Prom 14B & 0.174 & -0.095 & 0.070 & -0.079 & \pop{0.765} & 0.214 & -0.12 & 0.093 & -0.142 & \pop{0.921}\\
    Selene Mini & 0.107 & 0.005 & -0.153 & 0.011 & 0.404 & 0.103 & -0.016 & -0.202 & 0.038 & 0.561 \\
    Qwen2.5 7B & 0.138 & -0.106 & -0.103 & -0.166 & 0.538 & 0.178 & -0.147 & -0.088 & -0.207 & 0.718 \\
    Qwen2.5 72B & 0.160 & -0.037 & -0.097 & -0.096 & 0.272 & 0.220 & -0.050 & -0.102 & -0.111 & 0.412 \\
    GPT-4o mini  & 0.434 & -0.116 & -0.199 & -0.130 & 0.624 & 0.576 & -0.112 & -0.280 & -0.177 & 0.783 \\
    GPT-4o       & \pop{0.587} & 0.032 & -0.029 & 0.154 & \pop{0.698} & 0.722 & 0.105 & -0.055 & 0.218 & \pop{0.876} \\
    \bottomrule
\end{tblr}
    \caption{Rank correlations between automatic metrics and human annotations for the Spanish data. Largest three correlations for each criteria in \pop{bold}.}
    \label{tab:correlation-es}
\end{table*}

\end{document}